\documentclass{article} 
\usepackage{iclr2022_conference,times}
\usepackage[utf8]{inputenc} 
\usepackage{newunicodechar}
\usepackage[T1]{fontenc}    
\usepackage{hyperref}       
\usepackage{url}    
\usepackage{subfigure}
\usepackage{booktabs}       
\usepackage{amsfonts}       
\usepackage{nicefrac}       
\usepackage{microtype}      
\usepackage{bm}
\usepackage{amssymb}
\usepackage{newunicodechar}
\usepackage{amsmath}
\usepackage{amssymb}
\usepackage[lined,boxed,commentsnumbered, ruled,linesnumbered]{algorithm2e}
\usepackage{algorithmic}
\usepackage{multirow}
\usepackage{stmaryrd}
\usepackage{graphicx}
\newcommand{\bs}[1]{{\bm{#1}}}

\usepackage{amsmath,amsfonts,bm}

\def\eqref#1{equation~\ref{#1}}

\def\1{\bm{1}}

\DeclareMathAlphabet{\mathsfit}{\encodingdefault}{\sfdefault}{m}{sl}
\SetMathAlphabet{\mathsfit}{bold}{\encodingdefault}{\sfdefault}{bx}{n}

\usepackage{graphicx}
\usepackage{hyperref}
\usepackage{url}
\usepackage{natbib}

\title{FedGBF: An efficient vertical federated learning framework via gradient boosting and bagging}

\author{Yujin Han \\
Yale University\\
New Haven 06510, USA \\
\texttt{yujin.han@yale.edu} \\
\And
Pan Du \\
Renmin University of China\\
Beijing 100872, China\\
\texttt{dupan@ruc.edu.cn} \\
\And
Kai Yang \\
JD Technology Group\\
Beijing 100176, China\\
\texttt{yangkai188@jd.com}
}

\iclrfinalcopy 
\begin{document}

\maketitle
\begin{abstract}
Federated learning, conducive to solving data privacy and security problems,  has attracted increasing attention recently. However, the existing federated boosting model sequentially builds a decision tree model with the weak base learner, resulting in redundant boosting steps and high interactive communication costs. In contrast, the federated bagging model saves time by building multi-decision trees in parallel, but it suffers from performance loss. With the aim of obtaining an outstanding performance with less time cost, we propose a novel model in a vertically federated setting termed as Federated Gradient Boosting Forest (FedGBF). FedGBF simultaneously integrates the boosting and bagging's preponderance by  building the decision trees in parallel as a base learner for boosting. Subsequent to FedGBF, the problem of hyperparameters tuning is rising. Then we propose the Dynamic FedGBF, which dynamically changes each forest's parameters and thus reduces the complexity.  Finally, the experiments based on the benchmark datasets demonstrate the superiority of our method.
\end{abstract}

\section{Introduction }
With the increasingly widespread use of artificial intelligence, data brings great value in various industries. At the same time, data privacy and security are becoming increasingly important. Governments have introduced laws and regulations to protect data security, such as the General Data Protection Regulation \citep{GDPR} by European Union and California Consumer Privacy Act by California, United States. In China, data is also listed as one of the factors of production\footnote{\url{http://www.xinhuanet.com/politics/zywj/2020-04/09/c_1125834458.htm}}. The desire to collaborate different data owners to exploit the more value of data is blocked by data privacy and security concerns. To solve this problem, the concept of federated learning \citep{konevcny2016federated,yang2019federated,li2020federated} was proposed, making it possible to train machine learning models from multiple data sources without sharing data and privacy leakage.

Based on different data partitioning methods, federated learning can be categorized into three classes \citep{yang2019federated}. They are horizontal federated learning(HFL) \citep{konevcny2016federated}, vertical federated learning(VFL) \citep{gu2020federated,yang2019quasi}, and federated transfer learning \citep{liu2020secure}. Parties of HFL share the same feature space but differ in sample space. The VFL is characterized by different parties having different features of the same data space. And the federated transfer learning solves the problem where different parties differ both in sample space and feature space \citep{hardy2017private}.  In the business scenario, vertical federated learning is widely applicable and is considered as one of the effective solutions for encouraging enterprise-level data collaborations, as it enables both collaboratively training and privacy protection \citep{yang2019federated}.  In the application of VFL, especially in finance field, tree models are important because because it meets the user's requirements for model explainability \citep{bracke2019machine,bussmann2021explainable}.

As a type of tree model, Gradient Boosting Decision Tree (GBDT) is very popular and widely used in the risk control scenario. It provides an effective solution to the performance shortage of the decision tree model by integrating the multiple trees. GBDT can handle both classification and regression problems and  based on it, methods as XGBoost \citep{chen2016xgboost} and LightGBM \citep{ke2017lightgbm}, etc are proposed. SecureBoost \citep{cheng2021secureboost} is a privacy-preserving distributed machine learning which combines GBDT and federated learning.

SecureBoost makes it possible for the active party and passive party to access each other's data information without data sharing, ensuring the data security and privacy. However, SecureBoost model has the disadvantage of not being parallel and it only can be trained sequentially. In practice, many decision tree models are trained sequentially to guarantee a good performance which lead to inefficient modelling. Our local experiment results also support the view that SecureBoost is time consuming \ref{Time vs}.

To solve this problem, we propose a new method, FedGBF, which combines bagging and boosting method based on decision trees. FedGBF builds multiple decision trees in parallel at each boosting round and combine the results of each decision tree as the output of each layer. Because we apply the bagging idea at each layer, the performance of each forest layer in FedGBF outperform the single decision tree in SecureBoost thus less boosting round is required to achieve the similar performance \citep{breiman2001random,liu2012new}. In addition, we further propose the Dynamic FedGBF in order to reduce the complexity of FedGBF and avoid redundancy in model structure. In this paper, we have three main contributions.

\begin{itemize}

\item[$\bullet$] We propose a new vertical federated learning model, FedGBF, which combines bagging and boosting methods. The combination of these two methods allows us to obtain models with similar performance and less training times without large rounds of gradient boosting. 
\item[$\bullet$] Based on FedGBF, we propose an improved version of FedGBF. It’s Dynamic FedGBF, which dynamically adjust parameters to avoid redundancy and and reduces complexity.
\item[$\bullet$] We evaluate the dynamic FedGBF model on several VFL benchmark datasets and demonstrate that our approach significantly outperforms the existing baseline model SecureBoost in terms of efficiency. Dynamic FedGBF is much more efficiently than the baseline model SecureBoost while maintaining almost the same performance.
\end{itemize}

\section{Related Work}
\label{sec:relatedwork}

In the related work section, we mainly focus on three federated algorithms. All of them are the application of tree models to federated learning. They are Federated Forest, SecureGBM and SecureBoost.

\subsection{Federated Forest}
Federated Forest \citep{liu2020federated} is a CART tree and bagging based privacy-preserving machine learning model. The detailed steps of the algorithm are as follows.

\textbf{Step 1} Firstly, the active party randomly selects a subset of features and samples from the entire data and notify each passive party the selected features and sample IDs privately.

\textbf{Step 2} Then, each passive party calculates the local optimal split feature and corresponding impurity value and send this encrypted information to the active party. By comparing the purity improvement values of all alternative features, the active party can get the global best feature for splitting.

\textbf{Step 3} The active party notifies the passive party with the best splitting feature. The passive party will split the samples into left-right space and share the divided IDs information to the active party.

\textbf{Step 4} By step2 and step3, the decision tree can be built recursively. Repeat the above steps until the stop condition is reached.

\subsection{SecureGBM}

SeureGBM (\citep{feng2019securegbm}) is secure Multi-Party Gradient Boosting, which built-up with a multi-party computation model based on semi-homomorphic encryption \citep{bendlin2011semi}. SecureGBM needs to alignment the data of active party and passive party at first. Researchers usually adopt the private sets intersection algorithms \citep{huang2012private} to achieve this goal. Secondly, as the gradient boosting method, SecureGBM trains a decision tree to fit the negative direction of the gradient. And similar as federated forest, passive parties also need calculate local feature splitting and send these information to the active party. Then, the active party can have the best split feature as the node added into the decision tree. Finally, repeat above steps until reach the stopping condition.

\begin{figure}
   \centering
   \label{fig:FedGBF}
   \includegraphics[width=0.6\columnwidth]{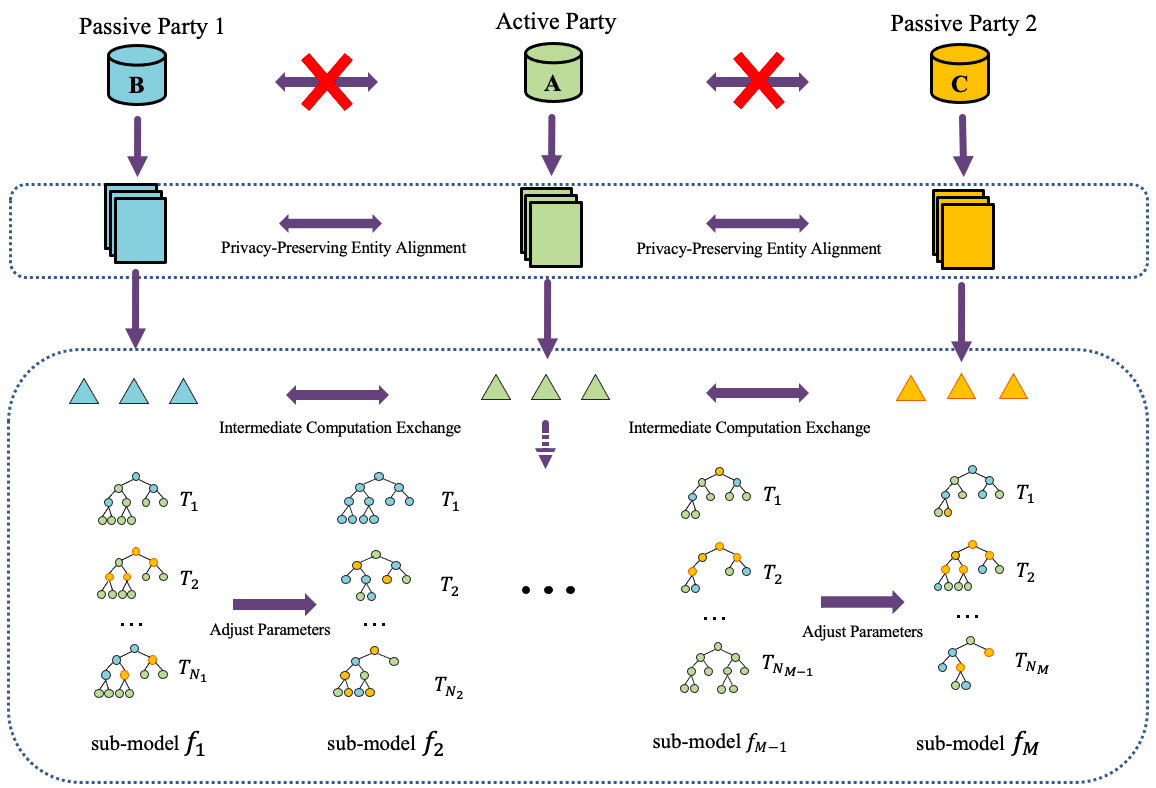}
   \caption{Illustration of the proposed FedGBF framework}
\end{figure}

\subsection{SecureBoost}

SecureBoost \citep{cheng2021secureboost} extends the machine learning model Xgboost within the federated learning framework. Steps of SecureBoost are as follows.

\textbf{Step 1} The active party with labels computes gradients of sample, and encrypts them to obtain $(⟨g_1⟩,⟨h_1⟩),…,(⟨g_n⟩,⟨h_n⟩$. Then active party sends these encrypted data to each passive party.

\textbf{Step 2} After binning, each passive party computes the first-order and second-order derivatives of samples through the formula $⟨G_L⟩=\sum_{i \in I_{L}}⟨g_i⟩$  and $⟨H_L⟩=\sum_{i \in I_{L}}⟨h_i⟩$ . Then each passive party sends these encrypted results to active party.

\textbf{Step 3} After decrypting intermediate results from passive parties, the active party has the $G_L ,G_R ,H_L ,H_R$ of each feature. Thus the indicator Gain in SecureBoost can be calculated by the formulation below.

\begin{equation}
\label{equ:1}
\begin{aligned}
L_{split}=  \frac{1}{2}[\frac{(\sum_{i \in I_{L}}g_i )^2}{(\sum_{i \in I_{L}}h_i+\lambda)}+\frac{(\sum_{i \in I_{L}}g_i )^2}{(\sum_{i \in I_{L}}h_i+\lambda)}-\frac{(\sum_{i \in I_{L}}g_i )^2}{(\sum_{i \in I_{L}}h_i+\lambda)}]-\gamma
\end{aligned}
\end{equation}

\textbf{Step 4} The active party compares the threshold $L_{split}$ for each feature and gets the best split feature of the present node. If the best feature is belong to passive parties, the active party need send the best feature information to corresponding passive party and get new sample space returned by the passive party. If the best feature is from the active party, the active party can split the sample space directly.

\textbf{Step 5} Repeat above steps until reach the status convergence or termination.

\section{Proposed Method}

\subsection{FedGBF: Federated Gradient Boosting Forest}

In this section, we propose Federated Gradient Boosting Forest (FedGBF), a novel tree ensemble model for federated learning by exploiting both the performance advantage of boosting and the efficiency advantage of bagging. In order to reduce the complexity of FedGBF, we then propose to dynamically change the important parameters, termed as Dynamic FedGBF.

\begin{algorithm}[H]
 \caption{FedGBF}
 \label{algorithm:FedGBF} 
\SetKwFor{Parfor}{for}{do}{in~parallel}
\SetKwInOut{Input}{Input}
\SetKwInOut{Output}{Output}
\Input{Training data $\bs{X}\in\mathbb{R}^{n\times d}$, Training labels $\bs{y}\in\mathbb{R}^{n}$, Boosting rounds $M$, Random Forest Size $N$, Sample sampling rate $\rho_{id}$, Feature sampling rate $\rho_{feat}$}
 \Output{FedGBF model $F$}
 Initialize $\hat{\bs{y}}^{(0)}=\bs{0}$\\
 \For{$m=1,\cdots,M$}{
 \Parfor{$j=1,\dots,N_m$}{
   According to $\rho_{id}$ and $\rho_{feat}$,  do the sample subsampling and feature subsampling of the data $\bs{X},\bs{y},\hat{\bs{y}}^{(m-1)}$, all parties can obtain $\bs{X}_m(j)\in\mathbb{R}^{n_m(j)\times d_m(j)},\bs{y}_m(j)\in\mathbb{R}^{n_m(j)},\hat{\bs{y}}^{(m-1)}(j)\in\mathbb{R}^{n_m(j)}$,
   where $n_m(j)=n\rho_{id}$ and
   $d_m(j)=d\rho_{feat}$\\\vspace{1em}

   Based on the sampled data, all parties construct $jth$ decision tree model $T_j=\textrm{GenerateTree}(\bs{X}_m(j),\bs{y}_m(j),\hat{\bs{y}}^{(m-1)}(j))$
 }
 Obtain $m_{th}$ Random Forest model $f_m(x_i)=g\Big(T_1(x_i),\dots,T_N(x_i)\Big)$.\\
 Update the predict value $\hat{\bs{y}}^{(m)}=\hat{\bs{y}}^{(m-1)}+f_m(\bs{X})$.
 }
After all sub-models are constructed, the FedGBF model is obtained. It is $F( \cdot)=\sum_{m=1}^{M}f_i(\cdot)$.

 \end{algorithm}
 
As with the SecureBoost and SecureGBM, our model is also inspired by GBDT. GBDT builds decision tree model sequentially. With the boosting method, GBDT can combine weak base learner to build a strong model. More specifically, there are $M$ sub-models $f_1,\cdots,f_M$. The $M$ sub-models are constructed to minimize the objective function:

\begin{equation}
\label{equ:2}
\begin{aligned}
{obj=\sum_{i=1}^n Loss(y_i,\sum_{m=1}^M f_m (x_i))+\sum_{m=1}^{M}(\Omega(f_m)))}
\end{aligned}
\end{equation}

Where $\Omega$ is the regularization term, which controls the complexity of the tree model. The construction of these $M$ sub-models uses the gradient boosting method. Specifically, in the $m-{th}$ step, the $m-1$ sub-models $f_1,…,f_{m-1}$ have been established, and the goal of $f_m$ is to minimize the loss function.
 
\begin{equation}
\label{equ:3}
\begin{aligned}
obj^m=\sum_{i=1}^n Loss(y_i,\hat{y}_{i}^{m-1} +f_m (x_i))+\Omega(f_m)+constant
\end{aligned}
\end{equation}

Where $\hat{y}^{m-1}=\sum_{j=1}^{m-1} f_j(x_i)$ is the output of the previously constructed model. 

To solve the inefficient problem of sequentially trained models like GBDT, we combine the random forest\citep{Friedman0Classification,2001Random} model and GBDT. In contrast to GBDT, the random forest constructs a large amount of decision trees in parallelization and combines these trees by bagging methods. Since random forest model makes use of bagging, it guarantees better modelling performance than a single decision tree. At the same time, the random forest has good generalization because it uses the idea of randomisation to make each decision tree in the forest different \citep{2001Random}. So Our method FedGBF takes the random forest model as a base learner of gradient boosting instead of a single decision tree. Similarly, we also randomly sample features and data to increase the generalizability of the model and also reduce the amount of computation workload. Based on the bagging method of random forest and the boosting method, FedGBF have the potential of achieving the similar performance as SecureBoost with much fewer boosting rounds.

There are main two steps of FedGBF. The first step is, after aligning the data under the privacy constraint, all parties need to collaboratively train a shared random forest model as the base learner. The second is the active party does the boosting gradient operation on the base learner. Algorithm 1 shows the process of training the FedGBF model \ref{algorithm:FedGBF}. Below, we give more detailed illustrations about each step.

The first step is training a random forest collaboratively. Random means participants are supposed to randomly select features and data for the model training. In our setting, the active party among the participants is main responsible for random selecting due to data privacy concerns. Specifically, first, participants should align the data under an encryption scheme by using the privacy-preserving protocol for inter-database intersections \citep{2004Privacy}. After this, the active party can know the code number of each passive party’s features and samples. Afterwards, the active party calculates the specific number of features and samples to be involved in the modelling, based on a pre-determined sample and feature sampling rate. The active party then informs each passive party of the selected feature IDs and sample IDs. It is important to note that, the active party does not know the specific feature information of each passive party and the active party only know the feature code and the corresponding passive party. Each passive party does not know how many features are selected in total. They only know what features they have are selected for this round of model training. Ensure that neither the active nor the passive party is fully aware of all information about the model can further protects all participants’ data security.

\begin{algorithm}[H]
\caption{GenerateTree}
\label{algorithm:generate_tree} 
\SetKwFor{Parfor}{for}{do}{in~parallel}
\SetKwInOut{Input}{Input}
\SetKwInOut{Output}{Out}
\Input{Training data $\bs{X}\in\mathbb{R}^{n\times d}$, Training labels $\bs{y}\in\mathbb{R}^{n}$, Current predict value $\hat{\bs{y}}^{(m-1)}\in\mathbb{R}^{n}$}
\Output{Tree model $T$}
After sampling data, a binning is performed for each feature dimension $k=1,\dots,d^p$ of $p_{th}$ party and the $L$ quantile points $S_k=\{s_{k1},\dots,s_{kL}\}$ of each feature can be obtained.\\
Active Party calculates the first order derivative of the loss function with respect to the current predicted value $g_i=\partial_{\hat{y}_i^{(m-1)}}Loss(y_i,\hat{y}_i^{(m-1)})$ and second order derivative $h_i=\partial_{\hat{y}_i^{(m-1)}}^2Loss(y_i,\hat{y}_i^{(m-1 }),i=1,\dots,n$.
Encrypt derivatives for transmission to each Passive Party party.\\
The initialized tree model $T$ contains only the root node Node0, and the corresponding sample set is the sampled sample set $I_0$. Initialize the set of nodes to be split $\mathcal{L}=\{Node0\}$.\\
\While{$\mathcal{L}$ is not Null}{
For each node $i\in\mathcal{L}$:\\
\textbf{Each party}
According to the binning threshold $S_k$, binning the data within the sample set $I_i$ corresponding to node $i$;\\
\textbf{Passive Party}
Using the public key, the first- and second-order derivative ciphertexts in  each bins corresponding to the samples of the features each Passive Party owns are summed and sent summation to Active Party party for decryption;\\
\textbf{Active Party}
Directly sum the first- and second-order derivatives in each bins of the features Active Party owns.\\
\textbf{Active Party}
The score for each split is calculated from the summation values of the first-order and second-order derivatives in each bins. According to the maximum score Score and corresponding splitting rule, the best split node can be obtained.\\
\uIf{Score$>\gamma$}{
According to the splitting rule, father node $i$ can be divided into two child node $i_L$ and $i_R$. The training data can also be divided into $I_L$ and $I_R$. Update the tree model$T_j$.\\
Add $i_L$ and $i_R$ into pending splitting set $\mathcal{L}$.}
\Else{
\textbf{Active Party}
Calculate the optimal weights $w_i$ of leaf node $i$
}
Remove node $i$ from the pending splitting set $\mathcal{L}$.
}
A tree model is constructed successfully.
\end{algorithm}

More specifically, the sampling process can be expressed by the following mathematical equation

\begin{equation}
\label{equ:4}
\begin{aligned}
(X_m(j),y_m(j),\hat{y}_{m-1}(j))=P_m (j)[XQ_m (j),y,\hat{y}^((m-1)]
\end{aligned}
\end{equation}

where $y_m (j)$is the corresponding data labels after sampling, $\hat{y}^{m-1} (j)$ is the prediction of the $m-1$ random forest models, $0<\rho_{id}=\frac{\tilde{n}}{n}\leq 1$ is the sample sampling rate and $0<\rho_{feat}=\frac{\tilde{d}}{d} \leq 1$ is the feature sampling rate. Denoting the sample sampling matrix by $P_m (j)\in R^{(\tilde{n}\times n)}$ and the feature sampling matrix by $Q_m (j) \in R^{(d \times \tilde{d} })$. After these operations, multiple decision trees are built in parallel.

The second step is the boosting operation. This step is the same as SecureBoost. After performing a gradient boosting, we can construct a new sub-model and repeat above steps until reach the stopping condition.


\subsection{Dynamic FedGBF}

FedGBF simultaneously integrates the boosting and bagging’s preponderance by building the decision trees in parallel as a base learner for gradient boosting. It reduces the time consuming by less boosting rounds and constructing random forests parallelly. However, FedGBF combining random forest and gradient boosting, has a very large model structure compared to SecureBoost. Complex models not only create large computational storage requirements, but also have the potential to cause overfitting that affects the generalisation of the model \cite{ying2019overview}. To further reduce the complexity, we propose the Dynamic FedGBF \ref{algorithm:Dynamic FedGBF} model.

\subsubsection{WHY DYNAMIC-FEDGBF IS NEEDED}
To better explain why we propose the Dynamic FedGBF model, we give the following three reasons.

\begin{itemize}

\item[$\bullet$] To reduce computational complexity. Although FedGBF takes advantage of the parallelism and reduce computational time, it still requires a significant number of decision trees to be compose the forest at each layer. These bring a huge computational overhead. If we dynamically adjust some parameters, such as the data sampling rate and tree numbers at each layer, this can, to a certain extent, reduce the complexity and computational overhead.

\item[$\bullet$] To mitigate overfitting and make the model more robust. Each forest in FedGBF incorporates the idea of bagging. Bagging, which reduces the over-reliance on a particular decision tree or a particular feature, achieves the effect of regularisation. When we further use different strategies to control parameters, each forest can be built from a much more different data set. Thus we can reduce the impact of anomalous data and make FedGBF more robust.

\item[$\bullet$] To increase Diversity. Just like why dropout was applies \citep{baldi2013understanding}, having different parameters in each layer can be regarded as having different forests. We can visualise this idea by assuming that $m_{th}$ forest in FedGBF has a large number of trees, but a small data size. And in contrast, the forest $j_{th}$ has a small number of trees but a large data size. So $m_{th}$ forest is wide but sparse and the forest $j_{th}$ is narrow but dense. Similar to nature, forests of FedGBF are different from each other, and the combination of these different forests with different characteristics can increase the diversity of species and makes the whole system more diversity.

\end{itemize}

We use both cosine and sine functions to dynamically control parameters we are interested in. The cosine annealing strategy\citep{loshchilov2017sgdr} is a very effective method in deep learning. Within the $i-th$ round, researchers decay the learning rate with a cosine annealing for each batch as follows.

\begin{equation}
\begin{aligned}
\eta_t=\eta_{min}^i+\frac{1}{2} (\eta_{max}^i-\eta_{min}^i )\times [1+cos(\frac{\pi T_{cur}}{T_i })]
\end{aligned}
\end{equation}

Where $\eta_{min}^i$ and $\eta_{max}^i$ are ranges for the learning rate, and $T_{cur}$ accounts for how many epochs have been performed since the last restart.

The use of cosine annealing allows the model to obtain different learning rates at different training epochs, avoiding the model falling into the local extremes. Inspired by this, a variant function of cosine annealing will also be used in FedGBF to adjust parameters. 

We use the cosine function to reduce the parameter values round by round and the sine function to increase the parameter values round by round. More specifically, in our paper, only two parameters, random forest data sampling rate and random forest decision tree number, are applied to control functions, while the rest of the parameters are fixed.  In order to reduce complexity and to avoid a single layer random forest with both a large number of trees and a large number of samples, our parameter tuning strategy is that the data sampling rate will gradually increase and the number of trees will decrease.

\begin{algorithm}[H]
\caption{Dynamic FedGBF}
\label{algorithm:Dynamic FedGBF} 
\SetKwFor{Parfor}{for}{do}{in~parallel}
\SetKwInOut{Input}{Input}
\SetKwInOut{Output}{Output}
\Input{Training data $\bs{X}\in\mathbb{R}^{n\times d}$, Training lables $\bs{y}\in\mathbb{R}^{n}$, Boosting rounds $M$, Minimum $S_{min}$, maximum $S_{max}$, and change speed $S_{k}$ of the sample, Minimum $t_{min}$, maximum $t_{max}$, and change speed $t_{k}$ of tree numbers, Feature sampling rate $\rho_{feat}$}
 \Output{FedGBF model $F$}
 Initialize $\hat{\bs{y}}^{(0)}=\bs{0}$\\
 \For{$m=1,\cdots,M$}{
 \Parfor{$j=1,\dots,N_m$}{
   According to $M$, $S_{min}$, $S_{max}$, $S_{k}$,$t_{min}$, $t_{max}$, $t_{k}$,$\rho_{feat}$, Active Party can calculate the numbers of sample, features and trees in this boosting and process sampling.

   Based on the sampled data,all parties construct $jth$ decision tree model $T_j=\textrm{GenerateTree}(\bs{X}_m(j),\bs{y}_m(j),\hat{\bs{y}}^{(m-1)}(j))$.
 }
 Obtain $mth$ Random Forest model $f_m(x_i)=g\Big(T_1(x_i),\dots,T_N(x_i)\Big)$.\\
 Update the predict value $\hat{\bs{y}}^{(m)}=\hat{\bs{y}}^{(m-1)}+f_m(\bs{X})$.
 }
After all sub-models are constructed, the Dynamic FedGBF model is obtained. It is $F(\cdot)=\sum_{m=1}^{M}f_i(\cdot)$.

 \end{algorithm}

\subsubsection{How to dynamically adjust parameters}

We can dynamically adjust parameters, such as tree numbers, data sampling rate and others, in each boosting round according to the formula below.

\begin{itemize}

\item[$\bullet$] Dynamic Increasing

\begin{equation}
\label{equ:6} 
\begin{aligned}
f_i(t)=\left\{
\begin{array}{l}
V_{min}+(V_{max}-V_{min}\cos{\frac{\pi(b_t-1)}{2k(b_T-1})} ,b_t\in[1,k(b_T-1)+1] \\
V_{min},b_t>k(b_T-1)+1\\
V_{max},b_T=1
\end{array}
\right.
\end{aligned}
\end{equation}


\item[$\bullet$] Dynamic Decaying

\begin{equation}
\label{equ:7} 
\begin{aligned}
f_i(t)=\left\{
\begin{array}{l}
V_{min}+(V_{max}-V_{min}\sin{\frac{\pi(b_t-1)}{2k(b_T-1})} ,b_t\in[1,k(b_T-1)+1]\\
V_{max},b_t>k(b_T-1)+1\\
V_{min},b_T=1
\end{array} \right.
\end{aligned}
\end{equation}
\end{itemize}

where $i$ implies that the $ith$ parameter. $V_min$ is the lower limit value of the parameter, $V_max$ is the upper limit value of the parameter, $b_t$ is the current round of boosting, $b_T$ is the total number of rounds of boosting, and $k$ is the parameter adjustment speed.

In particular, it is important to explain that the parameter $k$. Parameter $k$ controls the adjustment speed of the aimed one. For example, take the trees number parameter as an example, assume that the parameter is gradually decreased. We set that the total number of boosting is 11 rounds, the maximum number of decision trees allowed is 50 and the minimum number is 15. If $k=1$, this means that under the control of cosine function, the number of decision trees in each random forest layer will eventually decrease from 50 to 15 in the 11 boosting rounds. If $k=0.5$, the number of decision trees in each layer will decrease from 50 to 15 from the $1_{st}$ to the $6_{th}$ boosting round, and the number of decision trees will remain 15 from the $7_{th}$ to the $11_{th}$ boosting round.

\section{EXPERIMENTS}

\subsection{Dataset}
We evaluate our proposed method Dynamic FedGBF on two publicly available datasets and compare final results with the SecureBoost. In terms of performance, we report accuracy, Area under ROC curve (AUC) as well as F1 score. In terms of efficiency, we mainly compare the training and inference time of these two models.

\textbf{Give Me Some Credit Dataset} \footnote{\url{https://www.kaggle.com/c/GiveMeSomeCredit/data}}  is a credit score dataset that predicts the probability of a user defaulting. Banks can determine whether or not a loan should be granted based on the credit score. It contains a total of 150,000 instances and 10 attributes.

\textbf{Default of Credit Card Clients Dataset} \footnote{\url{https://www.kaggle.com/uciml/default-of-credit-card-clients-dataset}} is also a credit risk related dataset. It records information on credit card customers' default payments, demographic factors and more. It has a sample size of 30,000 and a feature number of 23.

There is no coherent vertical split for above data, so we refer to FATE \footnote{\url{https://github.com/FederatedAI/FATE}} for vertical data partition and divided the train data and test data by 7 to 3. 

\begin{table}[h!]\centering
\caption{Datasets}
\label{Table 1: Datasets}
\begin{tabular}{@{}ccccccc@{}}
\toprule
 & \multicolumn{2}{c}{Give Me Some Credit} & \multicolumn{2}{c}{Default of Credit Card Clients} \\ \cmidrule(l){2-5} 
 & active party & passive party & active party & passive party \\ \midrule
Samples & 150000 & 150000 & 30000 & 30000 \\
Dim & 5 & 5 & 13 & 10 \\
Classes & 2 & no label & 2 & no label \\ \bottomrule
\end{tabular}
\end{table}

\subsection{Experiment details}
In this section, we specify the details of our experiments. We adopt the idea of training the Dynamic FedGBF and SecureBoost models in a local rather than a federated environment and to compare the performance and runtime of these two local models. There is no any difference between the local and federated models in terms of model architecture, except no process of encryption, decryption and communication of intermediate results between participants locally. We will demonstrate theoretically and empirically that it is reasonable to 
evaluate our method through local modelling.

\subsubsection{Performance}

It is reasonable to use the same local model to estimate the performance of the federated model, including the AUC, ACC, and F1-score. Because there is no any approximation operations in the algorithm, these two models should be lossless constraint which means the loss of federated model over the training data is the same as the loss of local model which is built on the union of all data \citep{cheng2021secureboost,zhang2021secure}.

\subsubsection{Runtime}

We estimate the running time of both models using the following approach. At fist, fix the maximum depth $D$ and other parameters of a single decision tree, we train one decision tree on FATE using the all data and features and record the runtime as the unit time $T_{unit}$. Also, we record the time for data upload, data alignment and other pre-processing operations before training as $T_{0}$. Then, locally, we estimate the runtime $T_{single}$ for one decision tree with the same maximum depth of $D$ and other parameters based on different feature sampling rate and data sampling rate. Ultimately, according to $T_{single}$, we can obtain the estimated times for Dynamic FedGBF and SecureBoost.

\begin{equation}
\label{equ:8} 
\begin{aligned}
T_{single}=\alpha \beta T_{unit}
\end{aligned}
\end{equation}

where $\alpha$ is the data sampling rate and $\beta$ is the feature sampling rate.
 
\begin{equation}
\label{equ:9} 
\begin{aligned}
T^{L}_{F}=T_{0}+\sum_{i=1}^M T_{single,i}=T_{0}+\sum_{i=1}^M \alpha_{i}  \beta_{i} T_{unit}
\end{aligned}
\end{equation}

where $T_{G}$ is the estimated time of Dynamic FedGBF. At $ith$ round, the data sampling rate is $\alpha_{i}$ and the feature sampling rate is $\beta_{i}$. Decision trees at each layer of Dynamic FedGBF can be trained in parallel, but this is so ideal that it is difficult to satisfy multiple trees in parallel without congestion in practice. Thus we just regard equation \ref{equ:9} as the lower bound of the real runtime for Dynamic FedGBF.

\begin{equation}
\label{equ:10} 
\begin{aligned}
T^{U}_{F}=T_{0}+\sum_{i=1}^M N_{i}T_{single,i}=T_{0}+\sum_{i=1}^M \alpha_{i} \beta_{i} N_{i}  T_{unit}
\end{aligned}
\end{equation}

The equation \ref{equ:10} gives the upper bound of the real runtime for Dynamic FedGBF. $N_{i}$ is the number of decision trees in $ith$ round. It considers an extreme case where decision trees at the same layer of Dynamic FedGBF cannot be built in parallel and only modeled sequentially. The real runtime of Dynamic FedGBF$T_{G}$ should be between $T^{L}_{F}$ and $T^{U}_{F}$. 

Similarly, the estimated SecureBoost runtime formulation is following.

\begin{equation}
\label{equ:11} 
\begin{aligned}
T_{S}=T_{0}+\sum_{i=1}^M T_{single,i}=T_{0}+\sum_{i=1}^n \alpha_{i} \beta_{i} T_{unit}
\end{aligned}
\end{equation}

We set the learning rate for both models equals to 0.1 and the maximum depth of each decision tree is 3. The number of trees per layer of Dynamic FedGBF model will reduce from 5 to 2 with $k=1$ according to the equation \ref{equ:7} , and the data sampling rate $\alpha_{G}$ gradually increases from 0.1 to 0.3 with $k=1$ according to the equation \ref{equ:6}. Neither model samples for features which means $\beta_{G}=\beta_{S}=1$. To better evaluate the performance gap between these two models, we train the SecureBoost locally with $\alpha_{S}=1$.

As can be seen from the equations \ref{equ:8}\ref{equ:9}\ref{equ:10}\ref{equ:11}, we first train a single decision tree in a federated environment and use its modelling time as a unit time. And then, we train target models locally to estimate the runtime for the same model in a federated setting. The important assumption of the estimation method is that the time complexity of modelling a single decision tree is linearly related to the data size and number of features involved in the modelling. We demonstrate \ref{Time estimation method} that such an assumption is reasonable when the data size for modelling is large. What's more, to make our estimation method more convincing, based on the above parameter settings, we compare the estimated SecureBoost runtime with the real SecureBoost runtime. It can be noticed that the time estimation error is  less than 10\% and the estimation error rate decreases sharply with the number of boosting rounds increases \ref{Time vs}.

\subsection{Results}

\begin{figure}[h!]
\begin{minipage}[t]{0.3\linewidth}
\centering
\includegraphics[width=1.5in, height=1.5in]{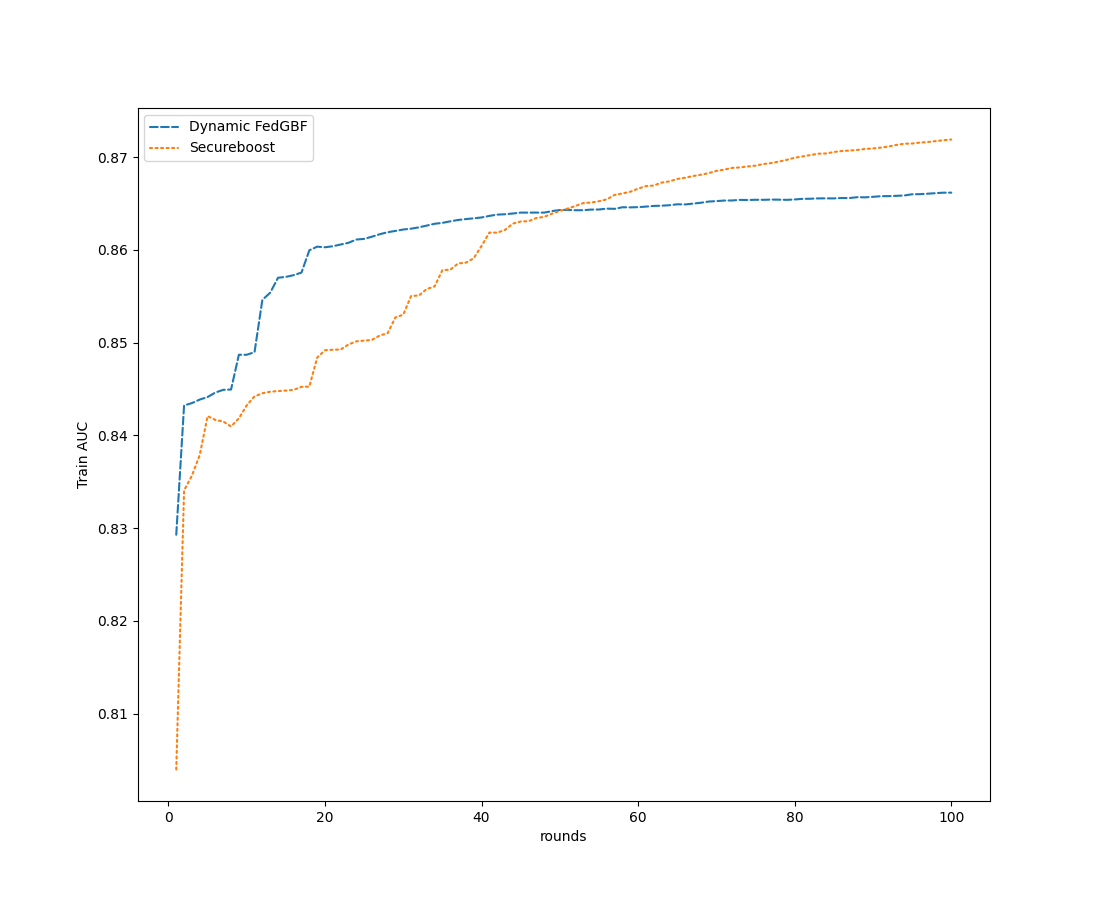}
\subfigure{(a) Train AUC}

\end{minipage}%
\begin{minipage}[t]{0.3\linewidth}
\centering
\includegraphics[width=1.5in, height=1.5in]{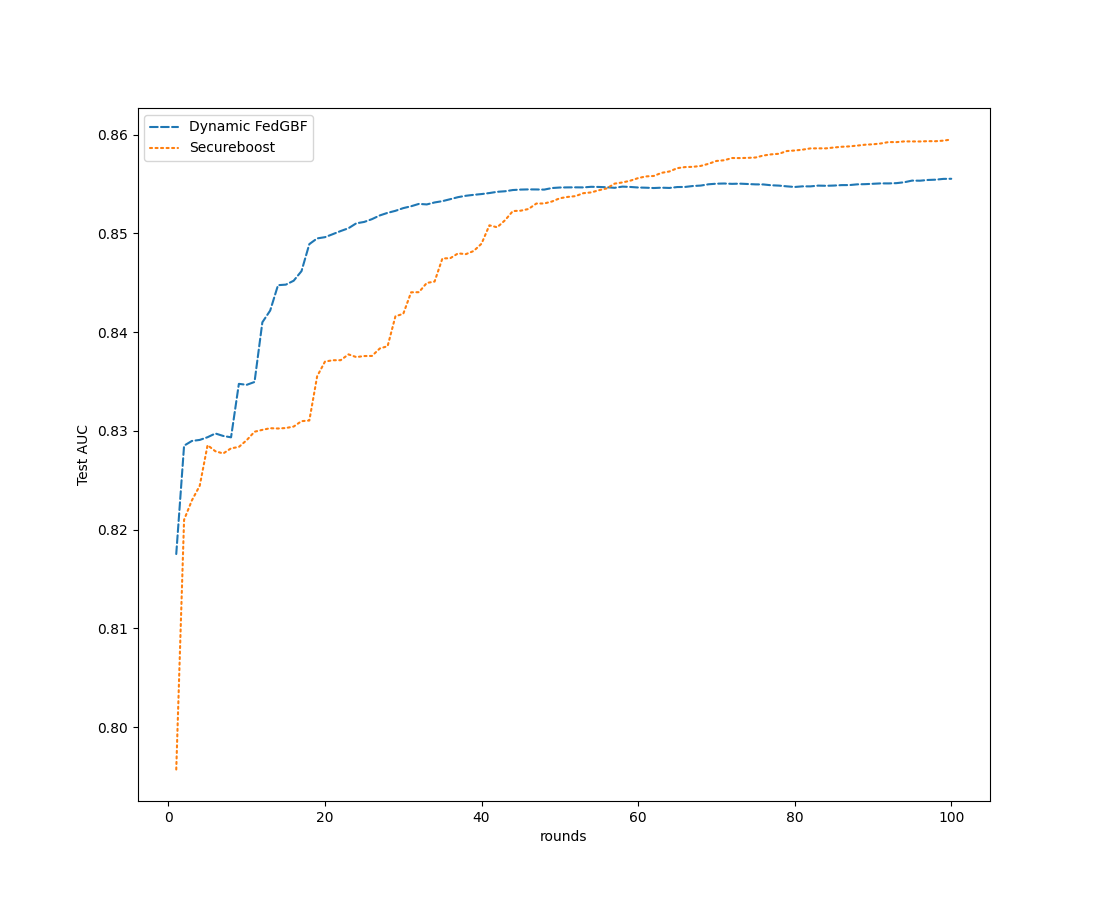}
\subfigure (b) Test AUC
\end{minipage}%
\begin{minipage}[t]{0.3\linewidth}
\centering
\includegraphics[width=1.5in, height=1.5in]{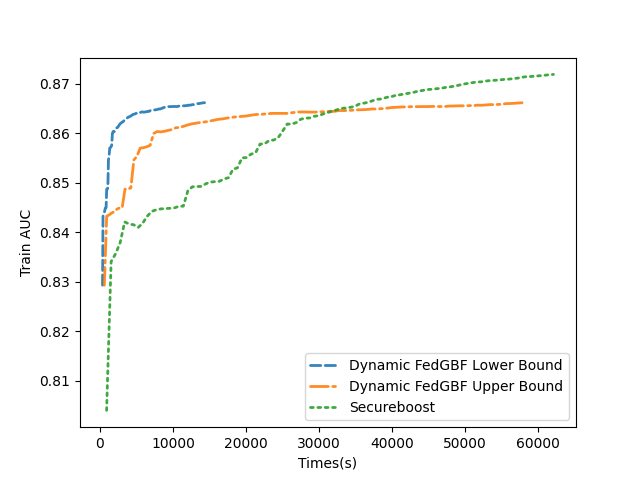}
\subfigure (c) Runtime vs. AUC
\end{minipage}
\caption{Performance and Runtime analysis on Give Me Some Credit Dataset}
\label{figure 1: giveme}
\end{figure}

\begin{figure}[h!]

\begin{minipage}[t]{0.3\linewidth}
\centering
\includegraphics[width=1.5in, height=1.5in]{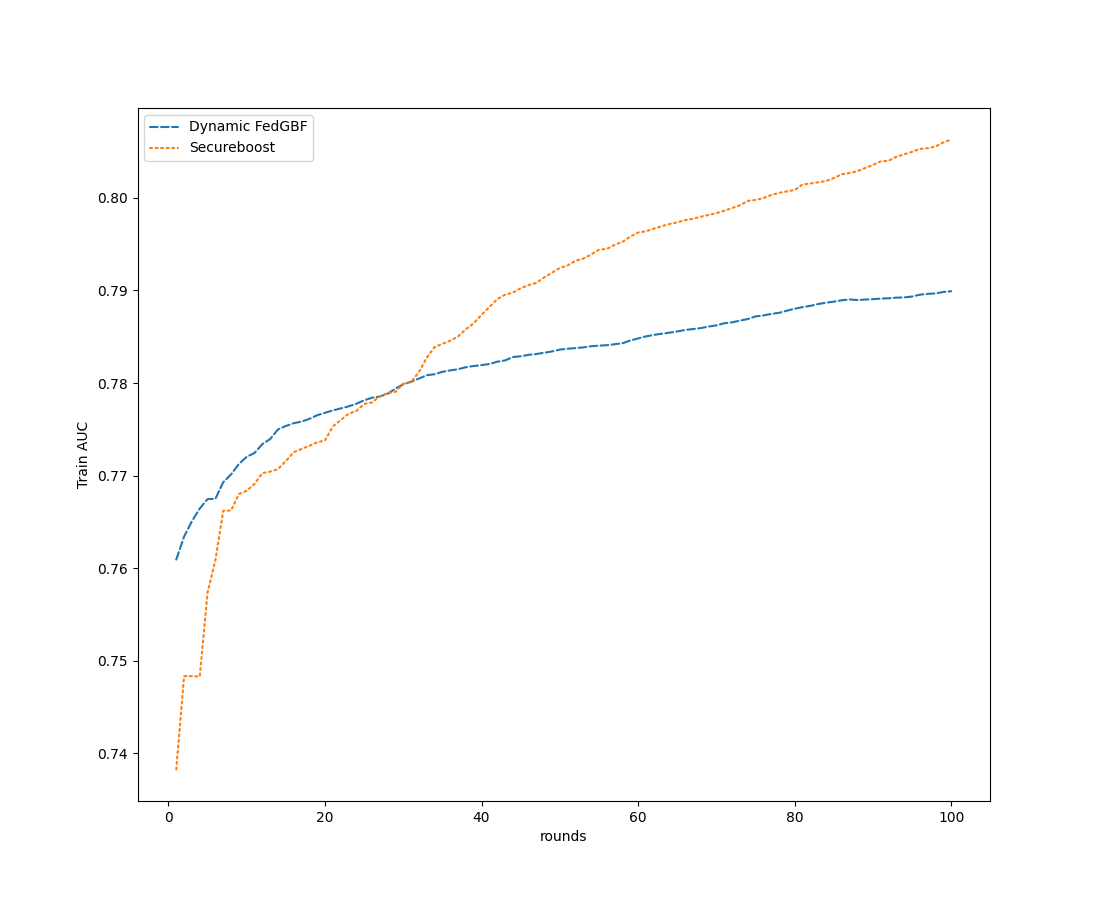}
\subfigure (a) Train AUC
\end{minipage}%
\begin{minipage}[t]{0.3\linewidth}
\centering
\includegraphics[width=1.5in, height=1.5in]{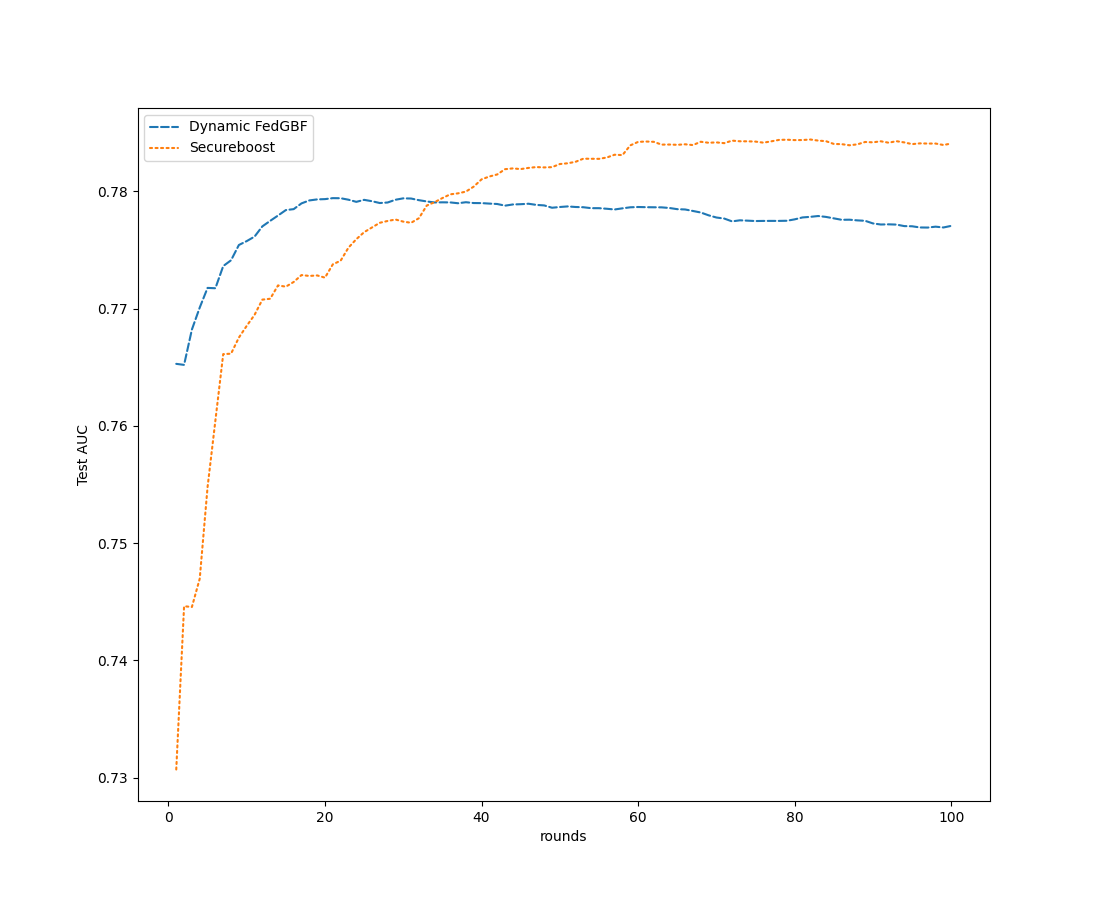}
\subfigure (b) Test AUC
\end{minipage}%
\begin{minipage}[t]{0.3\linewidth}
\centering
\includegraphics[width=1.5in, height=1.5in]{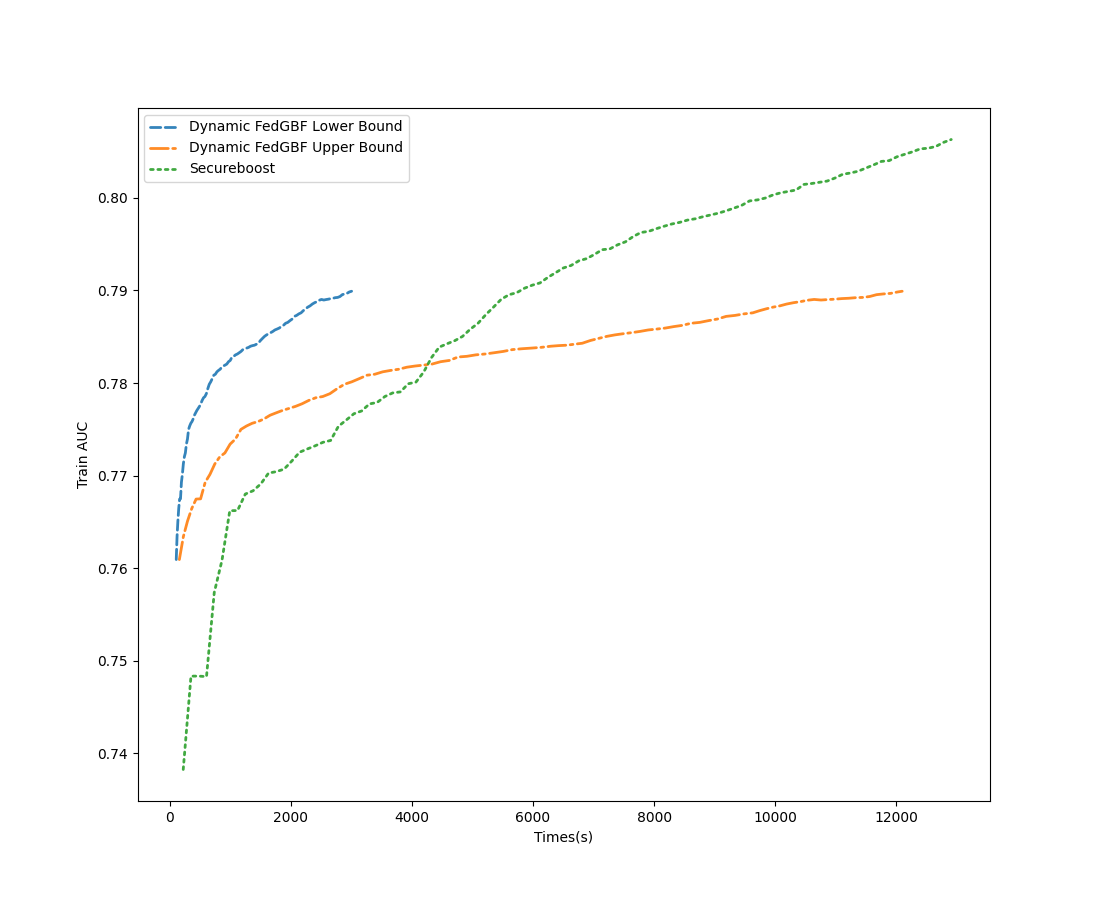}
\subfigure (c) Runtime vs. AUC
\end{minipage}
\caption{Performance and Runtime analysis on Default of Credit Card Clients Dataset}
\label{figure 2: credit}
\end{figure}

Table \ref{Table 2: Give Me Some Credit} and Table \ref{Table 3: Default of Credit Card Clients Dataset} show the performance results and the running time ($[T^{L}_{F},T^{U}_{F}] $ and $T_{S}$) of these two models for a given boosting rounds 20, 50 and 100. Figure \ref{figure 1: giveme} and figure \ref{figure 2: credit} provide more training details of the performance and time of these two models with the boosting rounds equals to 100.

\begin{table}[h!]
\caption{Performance and Runtime for Dynamic Fed-GBF vs. SecureBoost}
\label{Table 2: Give Me Some Credit}
\begin{tabular}{@{}ccccccc@{}}
\toprule
\multirow{2}{*}{Model} & \multirow{2}{*}{boosting rounds} & \multirow{2}{*}{Dataset} & \multicolumn{4}{c}{Give Me Some Credit Dataset} \\ \cmidrule(l){4-7} 
 &  &  & AUC & ACC & F1-Score & Estimated Time (s) \\ \midrule
\multirow{6}{*}{Dynamic FedGBF} & \multirow{2}{*}{20} & Train & 0.8591 & 0.9366 & 0.1543 & \multirow{2}{*}{[3086,11571]} \\%
 &  & Test & 0.8470 & 0.9352 & 0.1408 &  \\
 & \multirow{2}{*}{50} & Train & 0.8645  & 0.9381 & 0.2640 & \multirow{2}{*}{[7302,29104]} \\
 &  & Test & 0.8544 & 0.9371 & 0.2517 &  \\
 & \multirow{2}{*}{100} & Train & 0.8662 & 0.9382 & 0. 2733 & \multirow{2}{*}{[14328,58195]} \\
 &  & Test & 0.8555 & 0.9365 & 0.2514 &  \\
\multirow{6}{*}{SecureBoost} & \multirow{2}{*}{20} & Train & 0.8492 & 0.9366 & 0.216 & \multirow{2}{*}{12656} \\%
 &  & Test & 0.837 & 0.9359 & 0.2106 &  \\
 & \multirow{2}{*}{50} & Train & 0.8642 & 0.9377 & 0.2554 & \multirow{2}{*}{31198} \\
 &  & Test & 0.8536 & 0.9366 & 0.2432 &  \\
 & \multirow{2}{*}{100} & Train & 0.8719 & 0. 9386 & 0. 3064 & \multirow{2}{*}{62104} \\
 &  & Test & 0.8595 & 0.9371 & 0.2909 &  \\ \bottomrule
\end{tabular}
\end{table}

\begin{table}[h!]
\caption{Performance and Runtime for Dynamic Fed-GBF vs. SecureBoost}
\label{Table 3: Default of Credit Card Clients Dataset}
\begin{tabular}{@{}ccccccc@{}}
\toprule
\multirow{2}{*}{Model} & \multirow{2}{*}{boosting rounds} & \multirow{2}{*}{Dataset} & \multicolumn{4}{l}{Default of Credit Card Clients Dataset} \\ \cmidrule(l){4-7} 
 &  &  & AUC & ACC & F1-Score & Estimated Time (s) \\ \midrule
\multirow{6}{*}{Dynamic FedGBF} & \multirow{2}{*}{20} & Train & 0.7757 & 0.8199 & 0.4543 & \multirow{2}{*}{[674,2433]} \\
 &  & Test & 0.7771 & 0.8274 & 0.4537 &  \\
 & \multirow{2}{*}{50} & Train & 0.7844 & 0.8203 & 0.4829 & \multirow{2}{*}{[1548,6067]} \\
 &  & Test & 0.7797 & 0.8243 & 0.4746 &  \\
 & \multirow{2}{*}{100} & Train & 0.7899 & 0.8213 & 0.4900 & \multirow{2}{*}{[3004,12096]} \\
 &  & Test & 0.7771 & 0.8241 & 0.4774 &  \\
\multirow{6}{*}{SecureBoost} & \multirow{2}{*}{20} & Train & 0. 7738 & 0.8203 & 0.4741 & \multirow{2}{*}{2658} \\
 &  & Test & 0.7726 & 0.8257 & 0.4676 &  \\
 & \multirow{2}{*}{50} & Train & 0.7924 & 0. 8220 & 0.4792 & \multirow{2}{*}{6501} \\
 &  & Test & 0.7823 & 0.8257 & 0.4708 &  \\
 & \multirow{2}{*}{100} & Train & 0.8063 & 0. 8235 & 0. 4871 & \multirow{2}{*}{12906} \\
 &  & Test & 0.7841 & 0.8253 & 0.4707 &  \\ \bottomrule
\end{tabular}
\end{table}

As depicted in experimental results, we can draw the following conclusions. Firstly, in terms of model performance, Dynamic FedGBF and SecureBoost are very closed in terms of AUC, ACC and F1-Score. Dynamic FedGBF even performs better than SecureBoost in some given boosting rounds. This suggests that we can compensate for the performance loss of sampling by replacing a single decision tree in SecureBoost with the multiple decision trees in Dynamic FedGBF at each gradient boosting bound. Secondly, we further analyse the running time by comparing the runtime of SecureBoost and Dynamic FedGBF. In the ideal case, where decision trees are built in parallel at each layer of Dynamic FedGBF, Dynamic FedGBF's modelling time is only about 22\%-26\% of SecureBoost's, reducing modeling time by at least 74\%. In the worst case, where the decision trees in each layer of Dynamic FedGBF cannot be built in parallel and can only be modelled sequentially, the runtime of Dynamic FedGBF is also 6\% to 9\% shorter than Secureboost's. The reduction in Dynamic FedGBF running time is due to the dynamic data sampling, using up to 30\% of the original sample and at least 10\% of the original sample for each layer of the model. And the efficiency benefits of Dynamic FedGBF are even more greater if decision trees are ideally built in parallel.

The above experiments illustrate that Dynamic FedGBF can maintain good model performance while reducing training time significantly compared to SecureBoost.

\section{CONCLUSION}

In this paper, we propose a new federated learning model FedGBF. It addresses the problem of modelling inefficiencies of methods like SecureBoost which sequentially train models for gradient boosting. Based on FedGBF, we further adopt a dynamic adjustment strategy for parameters to avoid overfitting and structure redundancy. We compare the new approach Dynamic FedGBF and SecureBoost on two public datasets. Provided that these two model performances are very similar, Dynamic FedGBF, where decision trees at each layer can be built in parallel, takes only about 25\% of the runtime of Secureboost. And if decision trees at each layer are trained sequentially, Dynamic FedGBF's runtime is also shorter than Secureboost's. These experiment results reveal the truth that Dynamic FedGBF can indeed reduce the modelling time significantly without compromising the model performance compared to SecureBoost.

\bibliography{ref}
\bibliographystyle{iclr2022_conference}

\appendix
\section{Appendix}


\subsection{Time estimation method}
\label{Time estimation method}

The time complexity of decision tree construction without  re-sorting at each node is $O(mnlog_{2}(n))$ \citep{raschka38stat} where $m$ is the number of features and $n$ is sample size.

So the growth rate of the time to train a single decision tree is proportional to the growth rate of $f(n)=mnlog_{2}(n)$. Thus,

\begin{equation}
\label{equ:12} 
\begin{aligned}
\frac{T_{\alpha n,m}}{T_{n,m}}=\frac{m\times \alpha n\times log_{2}(\alpha n)}{m \times n\times log_{2}(n)}=\alpha+\frac{log_{2}(\alpha)}{log_{2}(n)}
\end{aligned}
\end{equation}

$\alpha$ in equation \ref{equ:12} denotes the data sampling rate and $\alpha \in (0,1]$. Compared to $\alpha$, $n$ is very large and $\frac{T_{\alpha n,m}}{T_{n,m}} \to \alpha$ when $\frac{\alpha}{n} \to  0$.

Regarding the relationship between running time and the number of features, we can draw similar conclusions.

\begin{equation}
\label{equ:13} 
\begin{aligned}
\frac{T_{n,\beta m}}{T_{n,m}}=\frac{\beta m\times n \times log_{2}( n)}{m \times n \times log_{2}(n)}=\beta
\end{aligned}
\end{equation}

where $\beta$ in equation \ref{equ:13} denotes the feature sampling rate.

\subsection{SecureBoost: Real Runtime vs. Estimated Runtime}

\begin{table}[h!]
\begin{tabular}{ccccc}
\hline
Data & Boosting Rounds & Estimated Time (s) & Real Time (s) & Error Rate \\ \hline
\multirow{3}{*}{\begin{tabular}[c]{@{}c@{}}Give Me Some \\  Credit\end{tabular}} & 20 & 12656 & 13818 & 8.41\% \\
 & 50 & 31198 & 32990 & 5.43\% \\ 
 & 100 & 62104 & 58538 & 6.09\% \\ 
\multirow{3}{*}{\begin{tabular}[c]{@{}c@{}}Default of Credit\\  Card Clients\end{tabular}} & 20 & 2658 & 2940 & 9.59\% \\
 & 50 & 6501 & 6888 & 5.62\% \\
 & 100 & 12906 & 13482 & 4.27\% \\ \hline
\end{tabular}
\end{table}

We calculate the error rate between the estimated running time and the true time by deflating the following formula.

\begin{equation}
\label{equ:14} 
\begin{aligned}
V_{Error Rate}=abs(1-\frac{V_{estimate}}{V_{real}})
\end{aligned}
\end{equation}

\label{Time vs}

\end{document}